\documentclass[sigconf,nonacm]{acmart}

\AtBeginDocument{%
  }

\usepackage{booktabs}
\usepackage{graphicx}
\usepackage[normalem]{ulem}
\useunder{\uline}{\ul}{}
\usepackage{multirow}
\usepackage{pifont}       
\usepackage{bbding}       
\usepackage{fontawesome}  

\begin{document}

\title{MM-Skin: Enhancing Dermatology Vision-Language Model with an Image-Text Dataset Derived from Textbooks}

\author{
    Wenqi Zeng$^{1}$, 
    Yuqi Sun$^{1}$, 
    Chenxi Ma$^{1}$, 
    Weimin Tan$^{1}$, 
    Bo Yan$^{1}$ \\
    $^1$Shanghai Key Laboratory of Intelligent Information Processing, School of Computer Science, Fudan University, Shanghai, China \\
}
\renewcommand{\shortauthors}{Zeng et al.}

\begin{abstract}
Medical vision-language models (VLMs) have shown promise as clinical assistants across various medical fields. However, specialized dermatology VLM capable of delivering professional and detailed diagnostic analysis remains underdeveloped, primarily due to less specialized text descriptions in current dermatology multimodal datasets. To address this issue, we propose MM-Skin, the first large-scale multimodal dermatology dataset that encompasses 3 imaging modalities, including clinical, dermoscopic, and pathological and nearly 10k high-quality image-text pairs collected from professional textbooks. In addition, we generate over 27k diverse, instruction-following vision question answering (VQA) samples (9$\times$ the size of current largest dermatology VQA dataset). Leveraging public datasets and MM-Skin, we developed SkinVL, a dermatology-specific VLM designed for precise and nuanced skin disease interpretation. Comprehensive benchmark evaluations of SkinVL on VQA, supervised fine-tuning (SFT) and zero-shot classification tasks across 8 datasets, reveal its exceptional performance for skin diseases in comparison to both general and medical VLM models. The introduction of MM-Skin and SkinVL offers a meaningful contribution to advancing the development of clinical dermatology VLM assistants.
\end{abstract}



\keywords{Dermatological Datasets, Vision-Language, Visual Question Answering}


\maketitle

\section{Introduction}

\begin{figure}
    \centering
    \includegraphics[width=\linewidth]{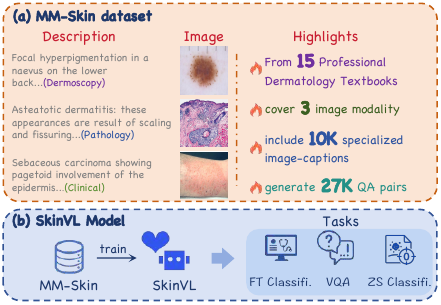}
    \caption{The proposed MM-Skin dataset and SkinVL model. MM-Skin is the first high-quality dermatology vision-language dataset featuring professional captions, multimodal images (clinical, dermoscopic, and pathological), and QA pairs. SkinVL, trained on MM-Skin, supports Visual Question Answering (VQA), supervised fine-tuning (SFT), and zero-shot (ZS) classification.}
    \label{fig:significant}
\end{figure}

Large Vision-Language Models (LVLMs) have shown strong capabilities in visual understanding and natural language generation \cite{bai2023qwenvlversatilevisionlanguagemodel,liu2023visual, ramesh2021zero, zhao2023easygen,feng2023towards, gu2024anomalygpt}. In healthcare, several LVLM s\cite{moor2023med,eslami2023pubmedclip,moon2022multi}, such as HuatuoGPT and LLaVA-Med\cite{zhang2023huatuogpt,li2023llava}, have been developed to assist clinical decision-making. In addition to general medical LVLMs, specialized models targeting specific medical domains have also emerged, such as XrayGPT \cite{thawakar2024xraygpt}, RepsNet \cite{tanwani2022repsnet}, and PathChat \cite{lu2024multimodal}, predominantly focusing on radiology or pathology. 

However, specialized dermatology LVLMs remain largely undeveloped. Recently, SkinGPT-4 \cite{zhou2024pre} was proposed as an innovative solution for interactive dermatology diagnosis. Despite its potential, SkinGPT-4 faces several constraints: 1) Training Datasets. SkinGPT-4's ability to generate specialized descriptions is limited by the narrow scope of Skincon \cite{daneshjou2022skincon} dataset, which includes only 48 clinical concepts, 1 image modality (clinical), and 3k images. While larger publicly available datasets are used for training, they provide only labels without image descriptions, further restricting the model’s ability to generate detailed responses; 2) Accessibility. Full model weights are not publicly available due to privacy concerns, hindering further development and accessibility; 3) Image Modality. SkinGPT-4 focuses only on clinical and dermoscopic images, excluding pathological images, which limits its comprehensive diagnostic capabilities. Given these challenges, developing a publicly available, specialized dermatology VLM is crucial for advancing the field.

The absence of dermatology-specific LVLMs is primarily due to the scarcity of public, high-quality dermatology image-text datasets, which are essential for generating grounded, interpretable, and instruction-following answers \cite{li2024gmai}. However, access to dermatology image descriptions is limited.  Unlike radiological images, which are typically paired with detailed reports\cite{zhang2024radgenome, johnson2019mimic}, clinical and dermoscopic images often lack written descriptions as part of the examination process, making it difficult to obtain expert descriptions. Additionally, most medical vision question answering(VQA) approaches \cite{nguyen2019overcoming,liu2021contrastive, chen2022multi, lin2023pmc} treat the task as classification or retrieval using predefined answer sets, limiting their ability to handle open-ended clinical questions. Currently, the only public dermatology VQA dataset, DermaVQA \cite{yim2024dermavqa}, is limited in size(3.5k images), modality (only clinical images), and answer quality (sourced from patient consultation websites with variable answer reliability). As a result, current dermatology LVLMs struggle with complex reasoning, highlighting the need for richer visual-language representations.

To address these challenges, we propose \textbf{MM-Skin} shown in Figure \ref{fig:significant}(a), a large-scale, high-quality, multimodal dataset specifically designed for dermatology. MM-Skin consists of nearly 10,000 image-text pairs covering 3 key imaging modalities: clinical photographs, dermoscopy, and pathology. All data are collected from 15 dermatology textbooks authored by experienced dermatologists, and each image is paired with long-form textual descriptions. To further support generative modeling, we leverage large language models (LLMs) to reformat these image-text pairs into over 27,000 instruction-following VQA samples. Compared to existing public dermatology datasets, MM-Skin offers a wealth of QA pairs and more detailed, professional, and diverse descriptions.

We developed \textbf{SkinVL} as shown in Figure \ref{fig:significant}(b), a domain-specific skin vision-language model, using MM-Skin and 9 public skin disease datasets, totaling 171.6k images. We evaluate the model on downstream tasks including VQA, supervised fine-tuning (SFT) and zero-shot diagnosis classification. Furthermore, we benchmark 7 representative LVLMs, including 5 general and 2 medical models, to systematically evaluate their performance. Our results highlight the importance of scalable and grounded medical datasets for improving the reasoning and generalization capabilities of LVLMs in healthcare. Our model has the advantage of generating more detailed, professional responses to a wider variety of questions.

\textbf{Our contributions are summarized as follows:}
\begin{itemize}
\item We propose MM-Skin, the first large-scale multimodal dermatology dataset with specialized long-text descriptions, including 10k image-text pairs and 27k QA pairs across 3 imaging modalities: clinical, dermoscopic, and pathological.
\item We develop SkinVL, the first publicly available dermatology-specific LVLM capable of generating detailed and clinically meaningful responses, tailored for precise and nuanced skin disease interpretation.
\item Comprehensive benchmarking of SkinVL on VQA, SFT, and zero-shot diagnosis classification tasks across 8 datasets demonstrates that SkinVL outperforms general and medical VLMs, demonstrating the effectiveness of MM-Skin for dermatology-specific visual understanding.
\end{itemize}

\begin{figure*}[ht]
    \centering
  \includegraphics[width=\linewidth]{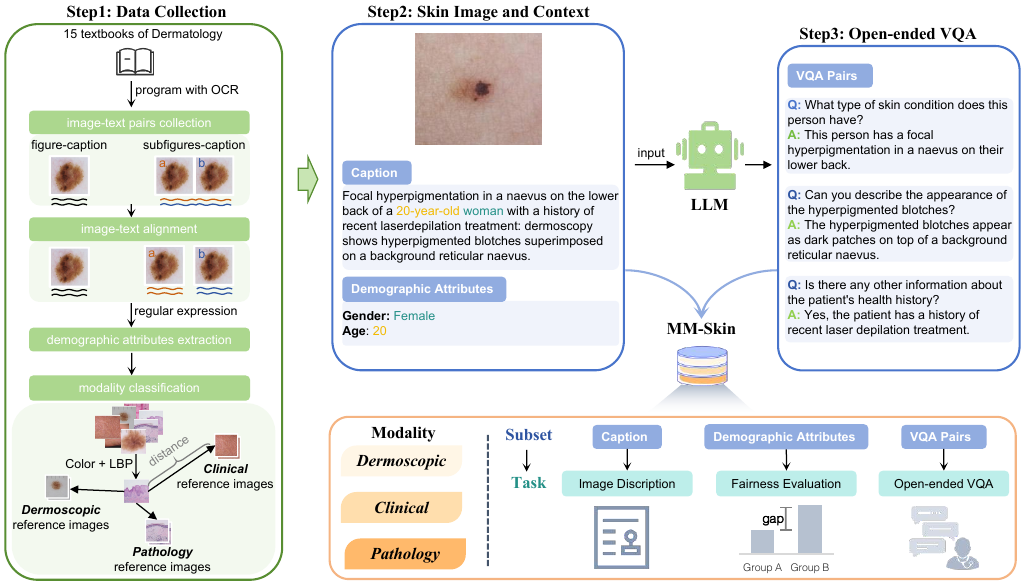}
  \caption{Illustration of the proposed pipeline for constructing MM-Skin, a dataset containing multimodal images, specialized captions, demographic attributes, and QA pairs, supporting multiple downstream tasks.}
  \label{fig:overview}
  \Description{A woman and a girl in white dresses sit in an open car.}
\end{figure*}

\begin{table*}[ht]
\centering
\caption{Comparison of public non-dermatology and dermatology medical VQA datasets with our proposed MM-Skin dataset.}
\label{tab:vqadatasets}
\resizebox{\textwidth}{!}{%
\begin{tabular}{@{}lccccccc@{}}
\hline
Domain &Dataset    & Modality                         & \# images & \# QA pairs & Answer Length                   & \# image-text pairs & \# demographic attributes \\ \hline
\multirow{6}{*}{Non-Dermatology} &PathVQA    & Pathology                        & 5K        & 32K         & 2.5                             &\ding{55}                   & \ding{55}                         \\
 &VQA-RAD    & Radiology                        & 0.3K      & 3.5K        & 1.6                             & \ding{55}                   & \ding{55}                         \\
 &SLAKE      & Radiology                        & 0.7K      & 14K         & -                               & \ding{55}                   & \ding{55}                         \\
 &VQA-Med    & Radiology                        & 5K        & 5K          & 1(74\%);2(2\%);  3(20\%);4(4\%) & \ding{51}                   & \ding{55}                         \\
 &PMC-VQA    & Mixture(80\%Radiology)           & 149K      & 227K        & -                               & \ding{55}                   & \ding{55}                         \\
 &OmniMedVQA & Mixture                          & 118K      & 128K        & -                               & \ding{55}                   & \ding{55}                         \\ \hline
\multirow{2}{*}{Dermatology} &DermaVQA   & Clinical                         & 3.4K      & 1.5K        & 11.9(66\%);  94.6(34\%)         & \ding{55}                   & \ding{55}                         \\
 &MM-Skin  & Clinical; Pathology; Dermoscopy & 11K       & 27K         & 21.67                           & \ding{51}                   & \ding{51}                         \\ \hline
\end{tabular}%
}
\end{table*}

\section{Related Work}

\subsection{Existing Medical VQA Datasets}

In recent years, an increasing number of datasets have been designed to advance medical VQA research. Table \ref{tab:vqadatasets} compares existing medical VQA datasets, which are crucial for medical LVLMs. Earlier datasets like VQA-RAD\cite{lau2018dataset}, VQA-Med\cite{ben2021overview}, SLAKE\cite{liu2021slake}, and Path-VQA\cite{he2020pathvqa} are limited by their small size (less than 20k) and their exclusive focus on radiology, which is insufficient for training high-performing models. Large-scale datasets such as PMC-VQA\cite{zhang2023pmc} and OmniMedVQA\cite{Hu_2024_CVPR} are constructed by generating question-answer pairs with the use of LLMs. Despite the quantity, images in PMC-VQA, 80\% of which pertain to radiology, were extracted from online papers, raising concerns about the quality and the diversity. Similarly, QA pairs of OmniMedVQA are derived from classification attributes that are not as descriptive or informative as captions, which may lead to the homogeneity of generated questions. Additionally, all these datasets suffer from limited text length, hindering their utility for more complex tasks like fine-grained image-text interactions and detailed medical instruction following QAs.

In the field of dermatology, long-text multimodal VQA datasets are notably scarce. Recently, the first public dermatology-specific VQA dataset DermaVQA was developed\cite{yim2024dermavqa}. However, it is constrained by inaccurate text descriptions, low-quality images collected from a mobile-based telemedicine platform, and a small sample size of around 3k images with a single modality (clinical images of patients), which restricts its broader clinical applications. In contrast, our dataset provides the first dermatology-focused long-text multimodal VQA dataset with nearly 10k high-quality image-text pairs and 27k question-answer pairs across dermoscopy, clinical images, and pathology modalities.

\subsection{Medical LVLMs}
The explosion of LVLMs has significantly advanced medical AI 
like medical consultation\cite{alkhaldi2024minigpt,liu2023visual} or disease diagnosis\cite{chen2024eyegpt,thawkar2023xraygpt}. The general medicine domain has witnessed the appearance of models like HuatuoGPT\cite{zhang2023huatuogpt}, LLaVA-Med\cite{li2023llava}, Med-Flamingo\cite{moor2023med}, PeFoMed\cite{liu2024pefomed} and Med-PaLm 2\cite{singhal2025toward}, while SkinGPT-4 has emerged as a dermatological specialist model\cite{zhou2024pre}. These models primarily use training data sourced from PubMed Central (PMC) and its variants, such as PMC-15M\cite{pmc15zhang2023large}, PMC-OA\cite{pmcoalin2023pmc}, or datasets generated by processing PMC-15M using LLMs, such as LLaVA-Med VQA\cite{li2023llava} and PubMedVision\cite{zhang2023huatuogpt}. LLaVA-Med, for instance, utilizes medical figure-caption pairs from PMC to scale the dataset and incorporates GPT-4 to transform contextual text into VQA tasks\cite{li2023llava}. However, LLaVA-Med still faces limitations in specialized medical fields like dermatology, as its training is primarily based on radiology and pathology, the two most common modalities. 

In dermatology, SkinGPT-4\cite{zhou2024pre}, an interactive dermatology diagnostic system, was trained on a large collection of public and private dermatology images with diagnostic labels. However, the domain-specific adaptation of SkinGPT-4 relies on an initial training set with only 3K images and 48 clinical concepts (Skincon\cite{daneshjou2022skincon}), limiting its ability to fully capture the diversity and heterogeneity of skin disease characteristics. Furthermore, this simplistic image-concept mapping may restrain the understanding and reasoning of the relationships among medical features displayed in the images. Hence, deploying LVLMs in real-world clinical scenarios is challenging, particularly for a patient's consultation other than diagnosis, highlighting the urgent need for specialized long-text training datasets in the dermatological domain.

\section{Methods}
\subsection{MM-Skin Construction}

\begin{figure*}[ht]
\centering
  \includegraphics[width=\linewidth]{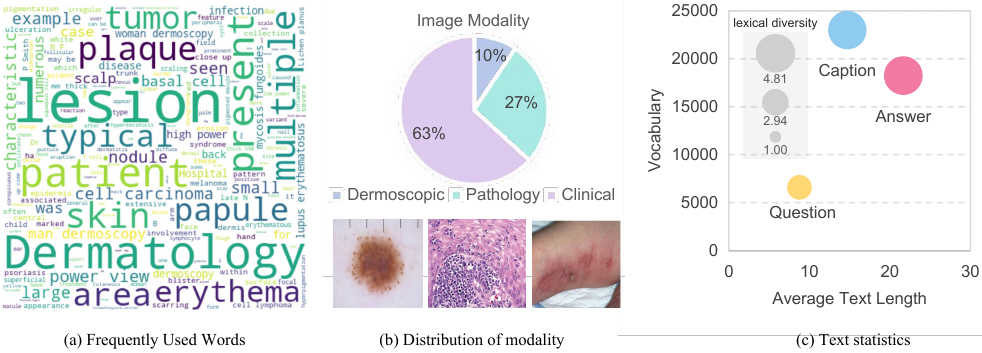}
  \caption{Statistical overview of the MM-Skin dataset. (a) Word cloud of caption texts, illustrating the diversity of dermatological terms. (b) Distribution of the three imaging modalities with representative examples. (c) Comparison of average text length, vocabulary size, and lexical diversity (Herdan’s C) for questions, answers, and captions.}
  \label{fig:data}
  \Description{A woman and a girl in white dresses sit in an open car.}
\end{figure*}
\subsubsection{Data Collection}
We will describe the construction of the MM-Skin dataset, with an overview in Figure \ref{fig:overview}. A semi-automatic pipeline with five steps was designed to create high-quality image-text pairs across dermoscopic, clinical, and pathology modalities.

\noindent\textbf{Image-Text Pair Collection}: First, image-text pairs are collected from 15 professional dermatology textbooks. Each image in the pair is ensured to have a resolution of at least 300 $\times$ 300 pixels. Following the procedure described in\cite{wu2024mm}, we utilize Adobe API and OCR techniques for raw image and text extraction. 

\noindent\textbf{Image-Text Alignment}: Some extracted images and texts do not align perfectly, as sub-figures share a single caption. We use regular expression matching to separate captions and align the image-text pairs accurately.

\noindent \textbf{Modality Classification}: Given that the images in the textbooks encompass multiple modalities, we categorize the images into dermoscopic, clinical, and pathology types. This classification is based on extracted features, such as the color histogram and local binary pattern (LBP). Let the extracted features for an image be denoted as \( \mathbf{x} \), which consists of the color histogram and LBP features. To classify an image, we extract features  \( \mathbf{x}_{\text{ref}_i} \) from reference images and construct a feature library. The test image \( \mathbf{x}_{\text{test}} \) is compared to each reference using a distance metric, and the image is assigned to the category corresponding to the reference image with the smallest distance:
\[d(\mathbf{x}_{\text{test}}, \mathbf{x}_{\text{ref}_i}) = \| \mathbf{x}_{\text{test}} - \mathbf{x}_{\text{ref}_i} \|\]
\[
\hat{y} = \arg \min_i \left( d(\mathbf{x}_{\text{test}}, \mathbf{x}_{\text{ref}_i}) \right)
\]
Finally, the classification is verified manually to ensure accuracy.



\noindent\textbf{Text Cleaning and Demographic Attributes Extraction}: Demographic attributes such as age and gender are extracted by matching relevant keywords in captions using regular expression techniques.

\noindent\textbf{Filtering and Processing}: Threads with images of genitalia, identifiable features (e.g., special tattoos), image annotations (e.g., drawn arrows), or full-face photos were removed. Detailed annotation guidelines are included in our data release.

\subsubsection{VQA Generation}
Although there are many Medical VQA datasets\cite{lau2018dataset,he2020pathvqa,liu2021slake} available, they all have two significant drawbacks that limit their practicality: (1) In the field of dermatology, open-source datasets for detailed question-and-answer interactions are notably scarce, especially those with professionally accurate text descriptions and large-scale data; and (2) Many current VQA datasets consist of simple closed-ended multiple-choice questions, while open-ended question-answering typically provides brief responses to straightforward queries, such as diagnostic questions, rather than generating long-form answers to diverse and complex inquiries. These limitations significantly restrict the flexibility and comprehensiveness of the datasets in addressing a broader range of user questions. Therefore, these issues highlight the need for a more extensive, long-text dataset like MM-Skin.

Large models are increasingly used to generate high-quality data, addressing data scarcity\cite{li2023llava,liu2024gemex,zhang2023huatuogpt}. As shown in Figure \ref{fig:overview}, we prepare to generate our VQA dataset after extracting image-caption pairs from professional dermatology textbooks. Here, we employ Llama-3.1\cite{touvron2023llama} as a generator to curate diverse instruction-following data with multi-round conversations about the provided skin images. Specifically, given an image caption, we design instructions in a prompt that asks the LLM to generate multi-round questions and answers in a tone as if it could "see" the image. Moreover, we also design specific rules (e.g., ensuring that no information not explicitly stated in the caption or context is introduced) to maintain the quality of the generated VQAs. The multi-round questions and answers are generated as follows:
\[\{q_1, a_1, q_2, a_2, \ldots, q_n, a_n\} = \text{LLM}(I, X)\]
where \(I\) is the image, \(X\) is the corresponding image description, and \(\{q_1, a_1, q_2, a_2, \ldots, q_n, a_n\}\) represents the sequence of multi-round questions and answers generated by the LLM. See Appendix B for the prompt examples. 

\begin{table}[t!]
\caption{Statistics of Image Type in MM-Skin.}
\label{tab: data}
\resizebox{0.48\textwidth}{!}{
\begin{tabular}{@{}lccccc@{}}
\toprule
Image Type  & \# Images & \# Image-text pairs & \# QA pairs & \# Gender & \# Age \\ \midrule
Dermoscopy & 1039      & 969                 & 2865        & 487       & 475    \\
Pathology   & 3016      & 2697                & 7834        & 14        & 6      \\
Clinical    & 6984      & 5703                & 16713       & 274       & 133    \\ \midrule
Overall     & 11039     & 9369                & 27412       & 775       & 614    \\ \bottomrule
\end{tabular}}
\end{table}

\begin{figure*}[!ht]
 \centering
  \includegraphics[width=\linewidth]{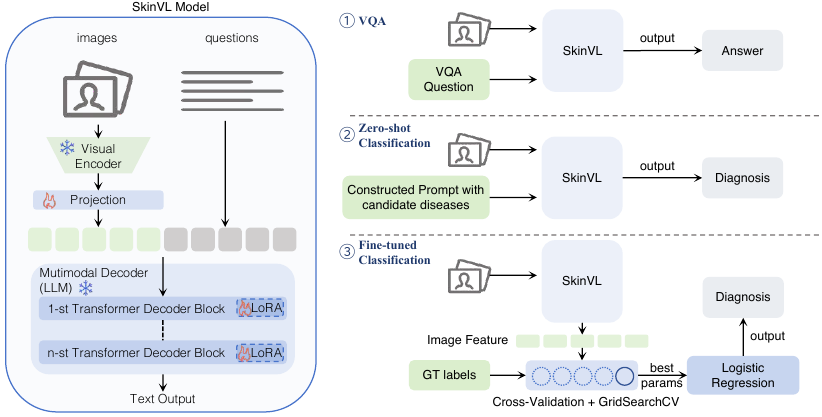}
  \caption{Overview of the SkinVL architecture, where the CLIP-ViT-L/14 visual encoder and language model decoder(LLaVA-Med) remain frozen, with only the visual projection layer and LoRA modules being updated. The figure also illustrates the evaluation procedures for visual question answering, supervised fine-tuning classification, and zero-shot classification. }
  \label{fig:model}
  \Description{A woman and a girl in white dresses sit in an open car.}   
\end{figure*}

\subsection{MM-Skin Statistics}

The current version of the MM-Skin dataset includes 11,039 images, of which 1,039 are dermoscopic, 3,016 are pathology, and 6,984 are clinical. The MM-Skin dataset can be divided into three subsets: MM-Skin-C (MM-Skin-Caption), MM-Skin-O (MM-Skin-OpenEnded), and MM-Skin-D (MM-Skin-Demographic), each corresponding to a specific downstream task. MM-Skin-C is used for image description, MM-Skin-O for open-ended VQA, and MM-Skin-D for fairness evaluation, which assesses whether the model performs equally well across different demographic groups (e.g., male and female patients). The number of corresponding professional text descriptions, QA pairs, and demographic sample information for each image is shown in Table \ref{tab: data}. A detailed statistical analysis of the data is provided in Figure \ref{fig:data}, focusing on aspects such as frequently used words, text length, and vocabulary size.

\noindent\textbf{Frequently Used Words}: To probe the diversity and coverage of image classes in MM-Skin, we generated a word cloud of the captions, shown in Figure \ref{fig:data}(a). Since the MM-Skin dataset is built based on comprehensive skin disease textbooks, it covers a wide range of disease categories, such as plaque, tumor papule, carcinoma, and erythema. Additionally, it includes words related to patient identity features, such as "child".

\noindent\textbf{Modality Distribution}: MM-Skin includes images from three skin imaging modalities. As shown in Figure \ref{fig:data}(b), clinical images constitute the majority (63\%), while dermoscopic images are the least represented (10\%). This diverse modality distribution offers broad coverage of skin disease types, which is essential for evaluating multimodal models in dermatology.

\noindent\textbf{Text Length and Vocabulary Size}: MM-Skin contains diverse textual descriptions, including disease diagnoses, lesion characteristics (e.g., color, shape, appearance), clinical manifestations, and post-treatment efficacy information. Figure \ref{fig:data}(c) displays the average text length, average vocabulary size, and lexical diversity across the three subsets. The average vocabulary size represents the total number of unique words across all the texts in the dataset, and lexical diversity is measured using Herdan's C, given by \( C = \frac{|V|}{\log N} \), where \( |V| \) is the number of unique words and \( N \) is the total number of words in the text. This metric helps assess the richness and diversity of the vocabulary used in the dataset.

\subsection{SkinVL Model}

\subsubsection{Fine-Tuned on MM-Skin} To validate the effectiveness of the MM-Skin dataset, specifically the auto-generated training set, we fine-tuned the LLaVA-Medv1-7B model\cite{li2023llava} on dermatology-specific multimodal datasets. We employed Low-Rank Adaptation (LoRA) instruction tuning to apply efficient parameter tuning on the LLaVA-Med backbone without retraining the entire model. This setup focuses on optimizing the parameters in the visual projector and the LoRA adapters, while keeping the vision encoder and the language model decoder frozen. All models are evaluated on three tasks: VQA, supervised fine-tuned diagnosis classification, and zero-shot diagnosis classification. The overall model structure is shown in Figure \ref{fig:model}. To investigate the impact of training data sources, we provide three variants of SkinVL, each trained on different datasets to assess how textual diversity and data quality affect performance:


\textbf{SkinVL-MM}: This model variant was trained solely on the MM-Skin dataset, which contains high-quality image-text pairs across clinical, dermoscopic, and pathological modalities, with a total of 21.8k images. Each image is paired with long specialized descriptions covering a comprehensive range of dermatological diseases.

\textbf{SkinVL-Pub}: Trained on the training sets of 9 public skin disease classification datasets, including DDI\cite{ddidaneshjou2022disparities}, HIBA\cite{hibaricci2023dataset}, BCN20000\cite{combalia2019bcn20000}, Fitzpatrick17k\cite{fitzpartrickgroh2021evaluating}, ISIC 2019\cite{isic2019codella2018skin, tschandl2018ham10000}, SCIN\cite{10.1001/jamanetworkopen.2024.46615}, PAD-UFES-20\cite{pacheco2020pad}, Patch16\cite{patch16kriegsmann2022deep}, and MSKCC\cite{MSKCCcodella2019skin}. The combined training set consists of 149k images in total. They contain diagnostic labels but do not provide QA-format supervision. To standardize with MM-Skin, we reformulated the diagnosis labels into question-answer pairs (e.g., “What is the diagnosis?” → “Melanoma”).

\textbf{SkinVL-PubMM}: This model was fine-tuned on a combined dataset of MM-Skin and reformatted public datasets, totaling 171.6k images, leveraging both specialized dermatology knowledge and a broader range of diagnostic labels.

\subsubsection{Downstream Tasks }The models were evaluated across 3 experimental tasks: VQA, supervised fine-tuning (SFT) classification, and zero-shot classification. The overall methodologies for these tasks are shown on the right side of Figure \ref{fig:model}.

\noindent \textbf{Visual Question Answering (VQA). }The model was provided with unseen images and related questions, allowing it to generate responses. The goal was to test its ability to generate accurate and contextually relevant answers based on visual content and assess how well it generalizes to new queries.

\noindent \textbf{Supervised Fine-tuned Classification. }To assess the generalization ability of visual representations, we extract image features from vision encoder and train a logistic regression classifier on downstream datasets using the corresponding ground-truth diagnostic labels. For each input image $I$, the visual feature is computed as:
\[
f(I) = P_{\text{mm}}(E_{\text{vision}}(I)) \in \mathbb{R}^d
\]
where $E_{\text{vision}}(\cdot)$ is the visual encoder and $P_{\text{mm}}(\cdot)$ is the multimodal projection layer. The logistic regression classifier is trained as:
\[
\hat{y} = \text{softmax}(W^\top f(I) + b)
\]
where $W \in \mathbb{R}^{d \times C}$, $b \in \mathbb{R}^C$, and $C$ is the number of disease categories. Hyperparameters are optimized using grid search over the regularization parameter $C \in \{10^{-5}, \ldots, 10^1\}$. We perform five-fold cross-validation using ROC-AUC as the selection metric. After finding the best hyperparameters, the classifier is retrained on the full training set and evaluated on validation and test splits.

\begin{figure}[h]
    \centering
    \includegraphics[width=\linewidth]{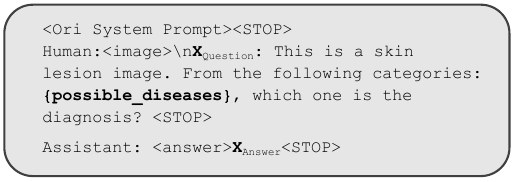}
    \caption{Input format for zero-shot classification.}
    \label{fig:promt}
    \Description{ss}
\end{figure}

\noindent \textbf{Zero-shot Classification. }To assess the model's diagnostic reasoning ability in a zero-shot setting, we designed a classification task without any training on the downstream data. Given an input image $I$, we constructed prompts $Q_C$  containing a list of candidate disease categories $\mathcal{C} = \{c_1, c_2, \dots, c_n\}$ and paired these with images, as shown in Figure \ref{fig:promt}.  The prompt $Q_C$ and image $I$ are jointly input into the model, which generates a textual answer $\hat{A}$. The prediction is then matched to one of the candidates in $\mathcal{C}$. This setup mimics real-world clinical applications where the model must infer diagnostic labels in the absence of task-specific supervision. It also evaluates the model’s capacity to perform grounded medical reasoning across previously unseen categories and datasets.

\section{Experiment}
\subsection{Experimental Setup}

\noindent\textbf{Evaluation Data. } We evaluated the model using test sets from the MM-Skin dataset and public datasets covering all three imaging modalities. We followed the official test set splits when available; otherwise, datasets were split 8:2 for training and testing. 
\begin{itemize}
    \item VQA: Evaluated on the MM-Skin test set, with 5.5k images.
    \item SFT classification: Evaluated on test sets from 8 public skin datasets across dermoscopic, clinical, and pathological modalities, with a total of 43k images. Specifically, the datasets include dermoscopic images from HAM10000 \cite{tschandl2018ham10000}, ISIC 2019 \cite{isic2019codella2018skin}, and HIBA \cite{hibaricci2023dataset}; clinical images from DDI \cite{ddidaneshjou2022disparities}, Fitzpatrick17k \cite{fitzpartrickgroh2021evaluating}, and PAD-UFES-20 \cite{pacheco2020pad}; and pathological images from Patch16 \cite{patch16kriegsmann2022deep}.
    \item  Zero-shot classification: we evaluated the model on test sets from Patch16 and additionally included HAM10000 \cite{tschandl2018ham10000} and Dermnet \cite{dermnet}, which were not seen during training, to test the model’s ability to generalize to unseen data under zero-shot conditions, totaling 33.5k images.
\end{itemize}

\noindent\textbf{LVLMs Benchmarks.} SkinGPT-4 was not included in our comparison due to the unavailability of its full model weights caused by privacy restrictions, yet we performed a qualitative comparison using its released partial weights (step-1) in the case study. Besides fine-tuning a task-oriented model, we evaluated 7 LVLMs, including 5 general-purpose models (doubao-1-5-vision-pro-32k, Gemini Flash2.0\cite{team2023gemini}, InternVL 2.5-8B\cite{chen2024expanding}, Qwen2.5-VL-7B\cite{bai2023qwenvlversatilevisionlanguagemodel}, LLaVA-v1.6-7B\cite{liu2024improved}) and 2 medical-domain models (LLaVA-Med-7B\cite{li2023llava}, HuatuoGPT-7B\cite{zhang2023huatuogpt}), to assess their capabilities in skin-related tasks.

\noindent\textbf{Training Details.} We fine-tune both the visual projection layers and LLM components of LLaVA-Med-v1.5 using LoRA, starting from the original model weights. Training is done on eight NVIDIA 3090 GPUs with a batch size of 16 per device. LoRA is applied with a rank of 128 and alpha of 64. The learning rate is 5e-5 with cosine decay, and gradient checkpointing is used to optimize memory. We also employ bf16 precision and Fully Sharded Data Parallel (FSDP) for better speed and scalability, with Deepspeed managing the training process and periodic model checkpoints.

\noindent\textbf{Evaluation Metrics.} We assessed the VQA models using BLEU-4, METEOR, ROUGE-L, BERTScore, and Recall to evaluate the linguistic accuracy and relevance of their responses. For classification tasks, we employed Accuracy, Precision, and Balanced Accuracy (BACC) to measure the models’ effectiveness in correctly identifying disease types across various modalities. These metrics provided a comprehensive evaluation of model performance.

\begin{table*}[]
\caption{VQA Performance of representative general LVLMs, medical LVLMs, and our SkinVL model on MM-Skin dataset.}
\label{tab:vqa}
\resizebox{\textwidth}{!}{%
\begin{tabular}{@{}l|ccccc|ccccc|ccccc@{}}
\toprule
    \multirow{2}{*}{Methods}              & \multicolumn{5}{c|}{Pathology}                                                                                                                      & \multicolumn{5}{c|}{Clinical}                                                                                                                       & \multicolumn{5}{c}{Dermoscopy}                                                                                                                    \\
                          & \multicolumn{1}{c}{Bleu-4} & \multicolumn{1}{c}{Meteor} & \multicolumn{1}{c}{Rouge-l} & \multicolumn{1}{c}{BertScore} & \multicolumn{1}{c|}{recall} & \multicolumn{1}{c}{Bleu-4} & \multicolumn{1}{c}{Meteor} & \multicolumn{1}{c}{Rouge-l} & \multicolumn{1}{c}{BertScore} & \multicolumn{1}{c|}{recall} & \multicolumn{1}{c}{Bleu-4} & \multicolumn{1}{c}{Meteor} & \multicolumn{1}{c}{Rouge-l} & \multicolumn{1}{c}{BertScore} & \multicolumn{1}{c}{recall} \\ \midrule
doubao-1-5-vision-pro-32k & 2.16                       & 22.45                      & 18.59                       & 56.20                         & 35.00                       & 2.96                       & 24.53                      & 20.23                       & 58.01                         & 36.58                       & 2.61                       & 24.03                      & 19.58                       & 57.76                         & 36.89                      \\
Gemini Flash2.0           & 3.66                       & 19.25                      & 17.93                       & 53.78                         & 26.97                       & 3.69                       & 19.30                      & 18.17                       & 54.80                         & 26.38                       & 3.87                       & 22.88                      & 19.33                       & 56.35                         & 31.27                      \\
InternVL 2.5-8B             & 3.22                       & 23.96                      & 20.68                       & 56.22                         & 37.27                       & 3.67                       & 25.22                      & 22.45                       & 58.27                         & 37.44                       & 3.50                       & 25.81                      & 22.56                       & 58.06                         & 38.90                      \\
Qwen2.5-VL-7B             & 1.97                       & 22.69                      & 16.88                       & 52.50                         & 42.70                       & 2.37                       & 24.31                      & 17.89                       & 54.65                         & 44.81                       & 1.89                       & 22.66                      & 16.26                       & 53.25                         & 47.47                      \\
LLaVA-v1.6-7B             & 2.14                       & 22.92                      & 17.60                       & 53.99                         & 47.25                       & 2.62                       & 24.91                      & 20.04                       & 56.19                         & 46.83                       & 2.32                       & 23.60                      & 19.11                       & 55.07                         & 46.25                      \\ \midrule
LLaVA-Med-7B              & 7.70                       & 32.24                      & 32.03                       & 63.57                         & 43.67                       & 7.07                       & 32.23                      & 31.88                       & 64.31                         & 42.92                       & 8.88                       & 33.74                      & 34.83                       & 66.60                         & 41.82                      \\
HuatuoGPT-7B              & 4.42                       & 29.03                      & 21.99                       & 58.42                         & 46.31                       & 4.14                       & 29.78                      & 21.69                       & 58.69                         & 47.15                       & 4.01                       & 29.21                      & 21.42                       & 58.49                         & 49.92                      \\ \midrule
SkinVL-MM                  & \textbf{22.04}             & \textbf{42.70}             & \textbf{44.09}              & \textbf{70.59}                & {\ul 48.20}                 & \textbf{23.39}             & \textbf{43.19}             & \textbf{44.32}              & \textbf{71.47}                & {\ul 47.66}                 & \textbf{21.61}             & \textbf{42.76}             & \textbf{44.70}              & \textbf{71.34}                & {\ul 47.51}                \\
SkinVL-Pub                 & 8.08                       & 31.85                      & 31.61                       & 62.24                         & 43.89                       & 6.94                       & 31.48                      & 30.39                       & 62.37                         & 43.66                       & 6.60                       & 29.99                      & 29.53                       & 60.94                         & 43.64                      \\
SkinVL-PubMM               & {\ul 16.56}                & {\ul 41.64}                & {\ul 39.66}                 & {\ul 68.27}                   & \textbf{51.29}              & {\ul 16.29}                & {\ul 41.66}                & {\ul 38.87}                 & {\ul 68.55}                   & \textbf{50.50}              & {\ul 13.48}                & {\ul 40.24}                & {\ul 37.20}                 & {\ul 66.89}                   & \textbf{52.26}             \\ \bottomrule
\end{tabular}%
}
\end{table*}

\begin{table*}[]
\caption{SFT Classification accuracy and recall of different models on skin disease datasets across three imaging modalities.}
\label{tab:ftclass}
\resizebox{\textwidth}{!}{%
\begin{tabular}{@{}lcccccccccccccccc@{}}
\toprule
\multirow{2}{*}{Methods} & \multicolumn{2}{c}{Patch16}     & \multicolumn{2}{c}{PAD-UFES-20}         & \multicolumn{2}{c}{ISIC 2019}        & \multicolumn{2}{c}{HIBA}        & \multicolumn{2}{c}{HAM10000}    & \multicolumn{2}{c}{Fitzpatrick} & \multicolumn{2}{c}{Dermnet}     & \multicolumn{2}{c}{DDI}         \\ \cmidrule(l){2-17} 
                         & ACC            & Recall         & ACC            & Recall         & ACC            & Recall         & ACC            & Recall         & ACC            & Recall         & ACC            & Recall         & ACC            & Recall         & ACC            & Recall         \\ \midrule
LLaVA-med                & {\ul 95.37}    & 94.27          & {\ul 73.48}    & {\ul 62.42}    & 75.93          & 56.77          & 80.43          & 80.34          & 78.42          & {\ul 62.13}    & 82.90          & 64.56          & 56.77          & 52.66          & \textbf{84.09} & 72.96          \\
SkinVL-Pub                & 95.23          & 94.15          & {\ul 73.48}    & 61.17          & 76.33          & {\ul 58.10}    & {\ul 81.35}    & {\ul 81.14}    & \textbf{79.29} & 60.19          & {\ul 83.18}    & {\ul 67.33}    & \textbf{57.62} & \textbf{53.88} & \textbf{84.09} & \textbf{74.88} \\
SkinVL-MM                 & 95.27          & \textbf{94.47} & 71.96          & 60.96          & {\ul 76.34}    & \textbf{58.21} & 79.20          & 79.11          & {\ul 79.15}    & 61.54          & \textbf{83.43} & \textbf{67.58} & {\ul 56.95}    & 52.42          & {\ul 83.33}    & {\ul 74.37}    \\
SkinVL-PubMM              & \textbf{95.63} & {\ul 94.43}    & \textbf{75.22} & \textbf{63.99} & \textbf{76.38} & 57.33          & \textbf{81.65} & \textbf{81.58} & 78.95          & \textbf{62.58} & 82.56          & 65.92          & 56.87          & {\ul 52.80}    & 82.58          & 71.94          \\ \bottomrule
\end{tabular}%
}
\end{table*}

\subsection{Evaluation on Visual Question Answering}
To ensure a fair comparison, all evaluated models, with the exception of doubao-1-5-vision-pro and Gemini Flash2.0, are based on 7B-LLMs. Specifically, LLaVA-v1.6 is based on Vicuna-7B-v1.5 \cite{zheng2023judging}, LLaVA-Med-v1.5 utilizes Mistral-7B-Instruct-v0.2\cite{jiang2310mistral}, and HuatuoGPT integrates the LLaVA-1.5 model architecture with LLaMA-3-8B\cite{zhang2023huatuogpt}. The configurations of all models were set according to their respective open-source codes. The results of these models are summarized in Table \ref{tab:vqa}, with the first five rows presenting the performance of general LVLMs, the next two rows representing medical LVLMs, and the final three rows showing the results of our fine-tuned version of LLaVA-Med based on different training sets.

Most general LVLMs perform poorly on MM-Skin. For instance, doubao-1-5-vision-pro-32k achieves only a Bleu-4 score of 2.16 in the pathology modality, reflecting its inability to generate relevant responses. LLaVA-v1.6-7B shows better recall but still scores low in Bleu-4 (2.14), indicating limited accuracy. Specialized models like LLaVA-Med-7B and HuatuoGPT-7B perform better, with LLaVA-Med-7B achieving 7.70 in Bleu-4 for pathology, and HuatuoGPT-7B scoring 4.42. The higher recall of HuatuoGPT-7B suggests a stronger understanding of medical images, highlighting the benefits of specialized training. The fine-tuned SkinVL-MM model, trained on MM-Skin, shows substantial improvement, achieving a Bleu-4 of 22.04 in pathology. It improves by nearly 10\% in Bleu-4 compared to InternVL 2.5-8B, the best general model for dermoscopic images, demonstrating the effectiveness of domain-specific fine-tuning. SkinVL-MM also excels in contextual relevance, with BertScores exceeding 70 across all modalities, outperforming both general and medical models. While SkinVL-MM trained solely on MM-Skin excels in natural language metrics, combining public data with MM-Skin (SkinVL-PubMM) enhances recall across all modalities. Dermoscopic images show the highest recall, likely due to their higher image quality, emphasizing the importance of image quality in model performance.

\begin{figure*}[ht!]
 \centering
  \includegraphics[width=\linewidth]{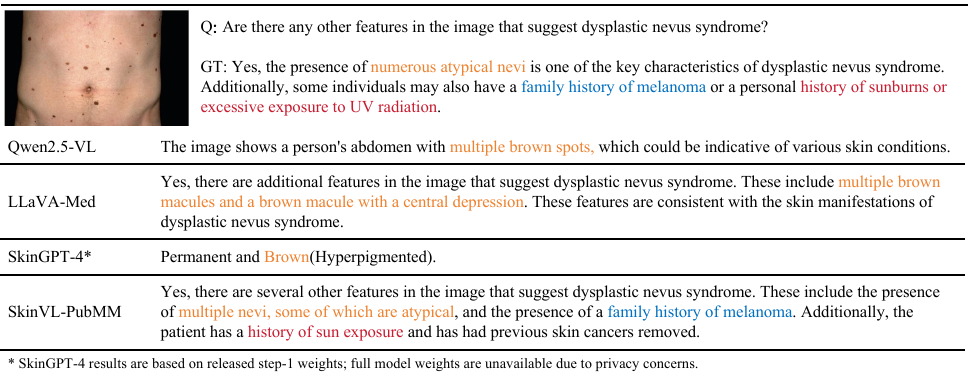}
  \caption{Qualitative comparison of medical visual question answering on diverse LVLMs using MM-Skin.}
  \label{fig:case}
  \Description{A woman and a girl in white dresses sit in an open car.}   
\end{figure*}

\subsection{Evaluation on SFT classification}

To assess the generalization ability of our fine-tuned models beyond the VQA task, we conducted a transfer evaluation on downstream classification. Specifically, we extracted image features from the trained multimodal models and trained a logistic regression classifier using the corresponding ground-truth diagnostic labels. This approach allows us to assess whether the visual representations learned under VQA-style supervision contain sufficient semantic information for skin disease classification, even without direct exposure to diagnostic labels. The evaluation was performed on eight publicly available skin disease datasets, covering three imaging modalities: pathology (Patch16), dermoscopy (ISIC 2019, HIBA, HAM10000), and clinical photography (PAD-UFES-20, Fitzpatrick, Dermnet, DDI). For datasets with official train-validation-test splits, we followed the original partitions. Otherwise, we adopted a 6:2:2 split. For Patch16, we simulated a low-resource setting by using only 10\% of the available training and validation data.

As shown in Table~\ref{tab:ftclass}, the model trained on both MM-Skin and public diagnostic datasets (SkinVL-PubMM) demonstrated the best overall performance, particularly on non-clinical images. It achieved the highest accuracy on Patch16 (95.63\%) and HIBA (81.65\%), and top recall on ISIC 2019 (57.33\%) and HAM10000 (62.58\%). Notably, SkinVL-PubMM surpassed SkinVL-Pub by +1.74\% on PAD-UFES-20 and by +0.32\% on ISIC 2019, highlighting the benefit of incorporating MM-Skin. Meanwhile, SkinVL-Pub marginally outperformed other models on clinical datasets, achieving the best performance on Dermnet (57.62\% ACC) and DDI (84.09\% ACC / 74.88\% Recall). These trends suggest that MM-Skin enhances feature learning for structured imaging modalities, while public datasets may better support generalization to more variable clinical photographs. Interestingly, the model trained exclusively on MM-Skin (SkinVL-MM), without access to any diagnostic labels during training, still achieved competitive results. Despite being optimized only for VQA using image-text supervision, it attained 76.34\% accuracy on ISIC 2019, 83.43\% on Fitzpatrick, and 79.15\% on HAM10000. Moreover, its recall on Patch16 (94.47\%) was nearly identical to that of SkinVL-PubMM (94.43\%), with only a 0.04\% difference. These results demonstrate that MM-Skin fosters the learning of transferable visual representations, making it a valuable resource not only for VQA tasks but also for broader applications in medical image understanding.

\begin{table}[ht!]
\caption{Zero-shot classification performance of representative LVLMs and SkinVL model.}
\label{tab:zsclass}
\resizebox{0.48\textwidth}{!}{%
\begin{tabular}{l|cc|cc|cc}
\toprule
\multirow{2}{*}{Methods}  & \multicolumn{2}{c|}{Patch16}          & \multicolumn{2}{c|}{PAD-UFES-20}        & \multicolumn{2}{c}{HAM10000}    \\
              & ACC                  & precision      & ACC            & precision      & ACC            & precision      \\ \midrule
Qwen2.5-VL-7B & 66.29                & 72.77          & \textbf{45.00} & {\ul 36.52}    & {\ul 39.58}    & {\ul 26.13}    \\
InternVL 2.5-8B & 60.97                & 43.00          & 32.83          & 21.76          & 30.97          & 17.02          \\
LLava-1.6     & 60.02                & 23.03          & 20.00          & 17.32          & 3.18           & 4.91           \\ \midrule
LLaVA-Med-7B     & {\ul 68.55} & {\ul 82.83}    & 32.39          & 7.76           & 5.49           & 1.43           \\
HuatuoGPT-7B     & 47.65                & 39.54          & {\ul 37.17}    & 29.07          & 19.79          & 18.65          \\ \midrule
SkinVL-MM      & 42.60                & 24.08          & 15.00          & 6.03           & 3.51 & 1.51 \\
SkinVL-Pub     & 68.44                & {\ul 84.22}    & 31.96          & 21.78          & 32.56          & 29.14          \\
SkinVL-PubMM   & \textbf{82.34}       & \textbf{86.65} & 35.65          & \textbf{41.88} & \textbf{51.29} & \textbf{37.37} \\ \bottomrule
\end{tabular}%
}
\end{table}

\subsection{Evaluation on Zero-shot classification}
To evaluate the generalization capabilities of various models in a more practical setting, we conducted a zero-shot classification task. In this scenario, models are required to make predictions without seeing any task-specific labeled training data. This setting closely reflects real-world deployment situations, where the model must infer the correct diagnosis from previously unseen cases. Evaluating zero-shot performance is thus essential for understanding the transferability and robustness of LVLMs. We performed this evaluation on three downstream datasets, each representing a different imaging modality: Patch16 (pathology), PAD-UFES-20 (clinical photography), and HAM10000 (dermoscopy). For each test image, we constructed a query containing all possible diagnosis candidates from the corresponding dataset and prompted the LVLM to select the most likely answer. We compared three general-purpose LVLMs, two medical-specific LVLMs, and three fine-tuned models based on our MM-Skin framework. The results are presented in Table~\ref{tab:zsclass}

Among all models, SkinVL-PubMM consistently achieved the best performance across datasets. On Patch16, it reached 82.34\% accuracy, outperforming LLaVA-Med by +13.79\%, and achieved the highest overall precision (86.65\%). On dermoscopic data (HAM10000), it remained the only model exceeding 50\% accuracy, while others lagged behind, highlighting the effectiveness of combining MM-Skin with public data during fine-tuning. SkinVL-Pub also performed well, with 68.44\% accuracy on Patch16, demonstrating the utility of public diagnostic data alone. In contrast, SkinVL-MM, which lacked diagnostic label supervision, struggled in the zero-shot setting, underscoring the importance of explicit label exposure. Among general-purpose models, Qwen2.5-VL-7B stood out with 66.29\% on Patch16, occasionally surpassing medical LVLMs. However, models like LLaVA-1.6 exhibited near-zero performance on HAM10000 (3.18\%), revealing the limitations of general LVLMs in specialized tasks. Overall, results demonstrate that fine-tuning with domain-specific multimodal data—particularly the combination of MM-Skin and public datasets—substantially enhances zero-shot classification, offering a practical path toward deploying LVLMs in real-world medical scenarios.

\subsection{Case Study}
Figure \ref{fig:case} presents a case comparing model responses to a query about dysplastic nevus syndrome. While Qwen2.5-VL provides a basic description and LLaVA-Med identifies some relevant features, SkinGPT-4, based on released available step-1 weights, uses simple terms and lacks the ability to adapt to diverse queries. In contrast, SkinVL-PubMM delivers a more comprehensive and accurate response, recognizing key visual features and integrating patient history, such as family history of melanoma and sun exposure. This demonstrates the value of fine-tuning with both MM-Skin and publicly available datasets to generate clinically grounded responses. Further case studies are available in the supplementary materials.


\section{Conclusion}
In this paper, we propose MM-Skin, the first open-access image-text dermatology dataset with 3 imaging modalities, and SkinVL, a specialized dermatology vision-language model. Our dataset improves model performance by providing high-quality, diverse image-text pairs, which play a pivotal role in advancing dermatology LVLMs. Further research into next-generation medical LVLMs with better reasoning capabilities will benefit from the improvement in the generation of detailed responses. We hope MM-Skin will contribute to the advancement of interactive AI systems in dermatology.

\newpage
\bibliographystyle{ACM-Reference-Format}
\bibliography{sample-base}

\appendix
\twocolumn[
  \begin{center}
    \Huge\bfseries Supplementary Materials: MM-Skin: Enhancing Dermatology Vision Language Model with an Authoritative Image-Text Dataset
    \vspace{0.5cm} 
  \end{center}
]

\begin{figure*}[hb]
    \centering
  \includegraphics[width=\linewidth]{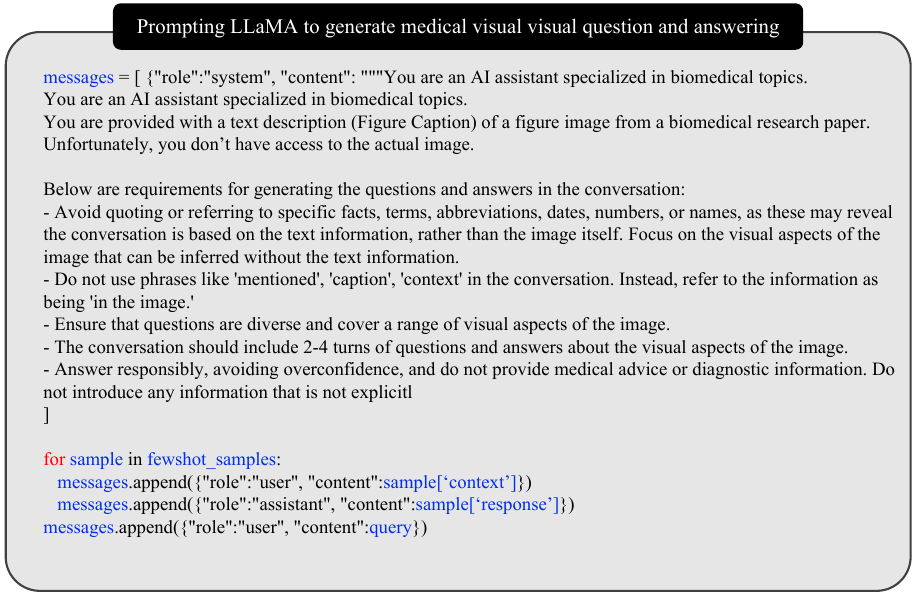}
  \caption{Illustration of the proposed pipeline for constructing MM-Skin, a dataset containing multimodal images, expert captions, demographic attributes, and QA pairs, supporting multiple downstream tasks.}
  \label{fig:promptllama}
  \Description{A woman and a girl in white dresses sit in an open car.}
\end{figure*}

\begin{figure*}[hb]
    \centering
  \includegraphics[width=\linewidth]{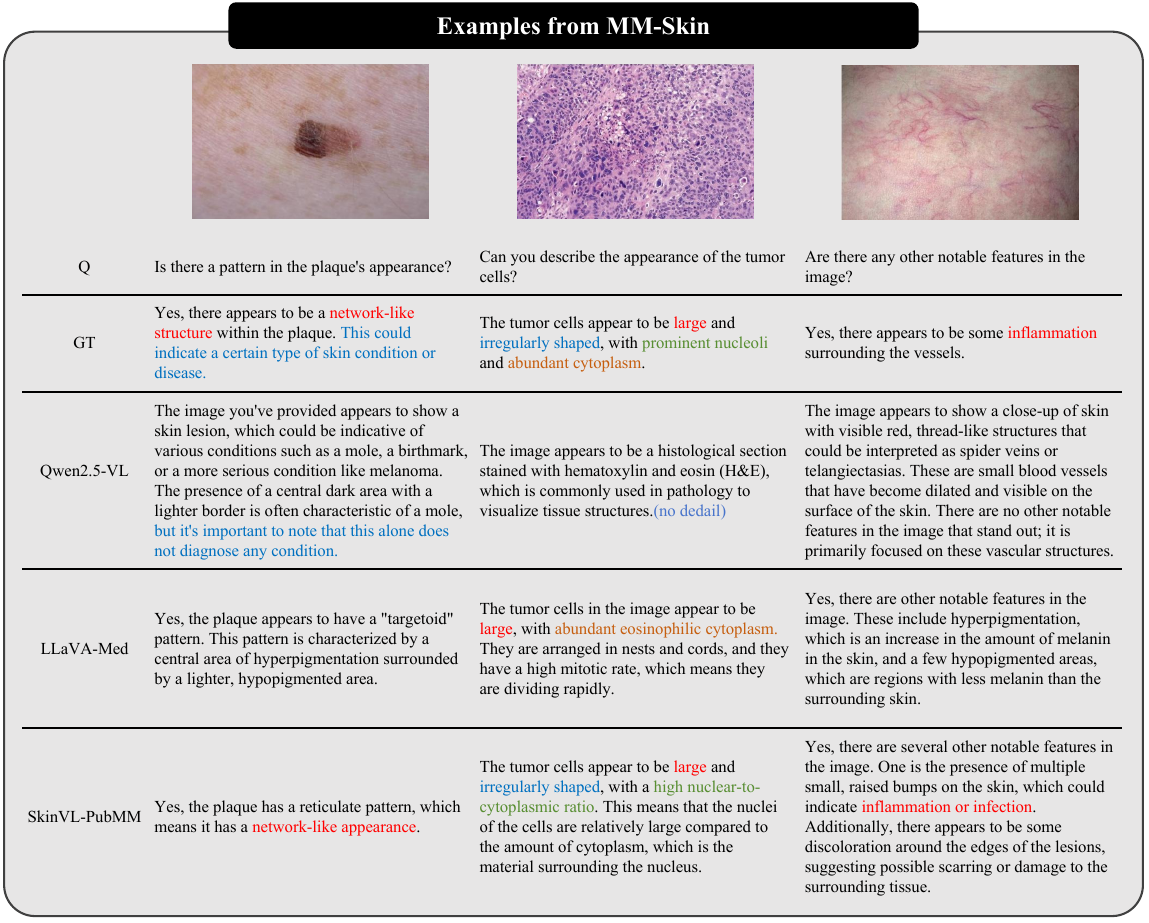}
  \caption{Example questions from MM-Skin answered by Qwen2.5-VL,  and LLaVA-Med our fine-tuned SkinVL-PubMM. These examples highlight the limitations of LVLMs like Qwen2.5-VL in visual reasoning, which can be effectively improved by fine-tuning with MM-Skin dataset.}
  \label{fig:caseappend}
  \Description{A woman and a girl in white dresses sit in an open car.}
\end{figure*}

\section{Prompt-Guided QA Generation}

Recent studies like MedPLIB and LLaVA-Med have shown the effectiveness of using Large Language Models (LLMs) for generating medical VQA data. Inspired by this, we use LLMs to create a high-quality dermatology-specific VQA dataset from our image-caption pairs. The dataset, consisting of skin images and expert-written captions, is expanded using a combination of LLM-based generation and manual selection to produce 27,000 diverse, clinically relevant QA pairs. We use LLaMA-3.1 as the QA generator, employing a prompt strategy designed to encourage diverse, image-inferable questions and simulate human-AI interaction. The process is detailed in Figure \ref{fig:promptllama}, ensuring consistency and relevance in the generated conversations.

\section{Case Study Extension}
To better understand the qualitative differences between general-purpose, medical-domain, and dermatology-specific LVLMs, we compare responses from Qwen2.5-VL, LLaVA-Med, and our fine-tuned SkinVL-PubMM on selected examples from MM-Skin (Figure \ref{fig:caseappend}). The comparison reveals distinct patterns in each model’s reasoning ability and domain alignment.

Qwen2.5-VL, as a general-purpose model, tends to rely on prior knowledge and surface-level cues. Its answers are often vague, speculative, or overly broad, lacking grounding in the actual visual evidence. For instance, when asked about lesion progression, it speculates about potential conditions (e.g., mole, melanoma) instead of describing the lesion’s visual changes over time. This suggests that without fine-tuning, general LVLMs struggle to adapt to domain-specific visual reasoning tasks. LLaVA-Med, trained on broader medical data, shows improved familiarity with medical terminology and structure. Its responses are more medically grounded but sometimes overconfident or overly specific, potentially introducing hallucinated details not directly supported by the input. For example, its mention of “crusted and ulcerated” lesions or “high mitotic rate” reflects a tendency to default to prototypical diagnostic language, which may not be appropriate for open-ended QA grounded in image descriptions. In contrast, SkinVL-PubMM demonstrates a more balanced and context-aware approach. Fine-tuned on MM-Skin, it produces responses that are not only clinically meaningful but also better aligned with the visual descriptions available, without introducing unsupported claims. Its answer regarding tumor cells, for example, emphasizes a high nuclear-to-cytoplasmic ratio and explains its significance in plain terms—highlighting both visual specificity and interpretability. Similarly, when asked about notable image features, it identifies signs like inflammation, raised bumps, and lesion edge discoloration, offering a more complete and nuanced interpretation than the other models. Overall, these comparisons suggest that general and even medical-pretrained LVLMs lack the necessary domain grounding and sensitivity to subtle visual details in dermatology. Fine-tuning on expert-annotated, multimodal data such as MM-Skin enables models like SkinVL-PubMM to generate more accurate, grounded, and interpretable responses in skin-related VQA tasks.

\section{Training and Evaluation Dataset Overview}

Table \ref{tab:dataset} presents the datasets used for training and evaluation of the SkinVL model. We used 10 datasets, including 9 public datasets and our MM-Skin dataset, with testing performed on the corresponding test sets. Additional zero-shot evaluation was conducted using the Dermnet and HAM10000 test sets. The datasets cover three image modalities: Pathology (53.62\%), Clinical (18.96\%), and Dermoscopy (21.14\%), with a similar distribution in the test set. Datasets like DDI, HIBA, and Patch16 were treated as binary classification tasks (Malignant/Non-Malignant), while others were multi-class, including ISIC (8 categories), MSKCC (21 categories), PAD (6 categories), Fitzpatrick17k (3 categories), HAM10000 (7 categories), and BCN20000 (12 categories). Dermnet, with the most categories (23), spans a range from Seborrheic Keratosis and Acne to Warts and Molluscum. These datasets ensure comprehensive evaluation.

\begin{table}[]
\caption{Training and evaluation dataset of SkinVL model}
\label{tab:dataset}
\resizebox{0.48\textwidth}{!}{%
\begin{tabular}{@{}lcccc@{}}
\toprule
                                  & Dataset        & Image Modality                        & \# Train & \# Test \\ \midrule
\multirow{11}{*}{Public Datasets} & SCIN           & Clinical                        & 10407    & -       \\
                                  & DDI            & Clinical                        & 525      & 133     \\
                                  & Fitzpatrick17k & Clinical                        & 12772    & 3194    \\
                                  & PAD            & Clinical                        & 1839     & 461     \\
                                  & Dermnet        & Clinical                        & -        & 4003    \\
                                  & HAM10000       & Dermoscopy                      & -        & 1512    \\
                                  & ISIC2019       & Dermoscopy                      & 17732    & 7601    \\
                                  & BCN20000       & Dermoscopy                      & 15157    & 3791    \\
                                  & HIBA           & Dermoscopy                      & 1309     & 328     \\
                                  & MSKCC          & Dermoscopy                      & 1037     & 260     \\
                                  & Patch16        & Pathology                       & 88972    & 28040   \\ \midrule
Ours                              & MM-Skin        & Clinical, Dermoscopy, Pathology & 21807    & 5452    \\ \midrule
\multicolumn{3}{c}{Total}                                                            & 171557   & 54775   \\ \bottomrule
\end{tabular}%
}
\end{table}

\end{document}